# A Universal Flexible Neuromorphic Tactile System with Multi-threshold Strategy

Jialin Liu

Diansheng Liao

ShiroYuki41706@gmail.com

shionri@outlook.com

## Abstract

Extremely increased unstructured data brought by the large-scale intelligent sensing devices application have big challenges not only in data storing and processing but also power consumption surging. Therefore, to improve energy efficiency and processing speed, a new generation system structure and construction strategy is necessary. Most biological nervous systems, especially the tactile system, have a good flexibility and data processing performance with low power usage. Inspired from this mechanism, to optimize the intelligent system, we report a universal fully flexible neuromorphic perception system with a strong compatibility and multi-threshold signal processing strategy by mimicking tactile nervous system. Peak signal accumulated from spike encoded sensor signal in front-end processing unit can be used for recognition task directly since the bionic synaptic plasticity. Compared with conventional systems, power consumption of our system significantly decreases about 1 order of magnitude in a same recognition task. What is more, the design of voltage-based matching circuit and multi-threshold processing circuit provide an excellent compatibility and multi-signal processing capability in our system. In feasibility verification, our system can output trend of different input signals (continuous signal and frequency signal etc.) accurately and have a high recognition accuracy of 90% in the symbol pattern and 90% in Morse code. These properties of our neuromorphic system show a great application potential in intelligent devices and bionic robots.

# 1. Introduction

With more application of artificial intelligence and intelligent robots in human daily life, data processing of intelligent perceptional devices increases rapidly day by day [1]. Large amounts of unstructured data cost more energy with severe transferring delay. In contrast, biological nervous system is energy-efficient, especially tactile perception system [2-5]. Tactile perception is a complicated system of nervous systems including different acceptors (thermoreceptors, pressure receptors, etc.) and neurons with various effectors connected to brain and medulla spinalis [6-8]. It is the most frequently used nervous system of organism for external environment interaction [9-12]. Compared to conventional processing architecture of sensor-ADC (Analog-to-Digital Converter)-memory-processor, biological system has advantage in high efficiency with lower power consumption an order of magnitude for same information processing at least [13-15]. A neuromorphic calculation paradigm based on bioinspired design for emulating the energy-efficient working of biological neural system has a diverse application prospect in biomedical [16, 17], electronic skin [18-21] and bionic robotics [22-24].

As core of intelligent perceptional device design, to the most extent, processing architecture has a determination of the power consumption and processing performance. Combining back-end neural network in conventional processing architecture with neuromorphic calculation paradigm can keep a high accuracy with low power usage [25-27]. However, sensor-ADC-memory-processor architecture in conventional intelligent perceptional devices can be optimized in data transmission and power consumption. Unstructured data in-situ preprocessing can reduce the latency and energy, so near-sensor calculation paradigm is proposed to suggest front-end processing unit be set near the sensors instead of ADC-processor unit [21-28]. By the bionic synaptic plasticity of neuromorphic front-end process unit (e.g. synaptic transistor and memristor) [29-31], according to spiking input format, valid information can be accumulated and converted to peak form (time interval information) while the invalid (background noise etc.) is too weak to process [21, 32-34]. Because of synaptic devices property [35-37] and connecting sensor unit to neuromorphic processing unit directly, input signal should be frequency information [16-24, 32]. To avoid the potential failure or even damage of neuromorphic processing unit caused by continuous changing signal collected from sensor unit, present neuromorphic system application scope is limited [18, 21]. What is more, only one threshold set in LIF (Leaky Integrate and Fire) model based neuromorphic processing unit cannot process the continuous-changing signal. Additionally, neuromorphic tactile systems have not yet achieved wide applications in those fields such as bionic robotics and E-skin since incomplete research in flexible neuromorphic tactile system large-scale manufacture [16, 23]. Compared to biological tactile system, current customized intelligent perceptional devices, limited by the design, have a single and specified functionality. To solve the problems above, proposing new materials, signal processing strategy and system construction strategy to build flexible and efficient general system for practical application is urgent.

This research proposed a universal flexible near-sensor neuromorphic tactile perception system based on modular design and multi-threshold signal processing strategy and gives a neuromorphic preprocessing solution of unstructured data in continuous & frequency signal. Pressure tactile perception as example, a type of 3D-printed based 3×3 flexible piezoresistive sensor array with high sensitivity (31.687 kPa$^{-1}$) and wide range (0-150 kPa) is prepared as sensing unit. The front-end processing unit is a type of 3×3 flexible synaptic transistor array made from P3HT film and [EMIM][TFSI] ion-gel. Benefit from spike-coding circuit between sensing unit and front-end processing unit, processing information in pulse is one of keys to achieve low power consumption. Meanwhile, sampling accumulated pulses filtered from the back-end unit can further reduce power consumption. Unlike single threshold in existing SNN model, multi-threshold signal processing circuit and processing algorithm increase the accuracy of pattern recognition (acc. 90%), Morse code recognition (acc. 90%) and the robustness of system by expanding input dimensions. Additionally, thanks to the voltage interface, this system has a different electrical sensor unit compatibility for multifunctionality. Power consumption decreases from 18.387 W to 1.755 W for same input information processing in contrast to conventional architecture.

## 2. Materials and methods

*2.1. Fabrication of 3D-printed Conductive Layer*

This conductive pressure sensitive layer is produced by digital light processing 3D printer (AUTOCERA-R). The composite conductive resin is mixed with UV curable resin (Stratasys, Agilus 30), MWCNT (radius/diameter: 3-15 nm, length: 15-30 um) and 1,6-hexanediol diacrylate (Aladdin, CAS 13048-33-4) with a mass mixing ratio of 100:0.5:3.5. Homogenizer (ZYE, ZYMC-200V) is used here to mix and homogenize composite resin by multi-period running mode (800 r/min(30s)-1500 r/min(60s)-1000 r/min(60s)) and dump it into printer cartridge as quick as possible. Exposure parameters of 3D printer are set as 75 mW/mm$^2$ and 1.5 s with 30 μm of each sliced layer. Scraping is needed before printing each layer. After printing, the printout should be flushed by isopropanol at least 3 times and shovel it from molding platform to curing box. After 6-minutes curing, we suggest that keep the material in nature light environment several days to improve performance such as less resistance and higher tensile strength.

*2.2. Fabrication of Flexible Sensor Array*

This 3×3 flexible sensor array is consisted by 3 parts: (i) Substrate is made from PI film (0.2mm). Before evaporating electrode, it needs to be processed by plasma cleaner (PVA TePla, IoN 40) (O$_2$, 100 W, 5 min). (ii) With electron beam evaporation deposition and masks with interdigital electrode shape, electrode layer is made from Cr(10 nm)/Au(90 nm) sediment on processed PI film. (iii) 9 parts (2.5 mm×2.5 mm each one) are cut from 3D-printed pressure sensitive layer and put on 3×3 array electrode. Finally, the pressure sensitive unit is fastened on interdigital electrode by PI film and tape. Notably, the area of pressure sensitive unit should be cut less than interdigital electrode.

*2.3. Characterization of Flexible Tactile sensors*

Sensor data are collected by materials testing systems (Instron, 5943 1SET). Under 1mA current by power source (Keithley), nano volt/micro-ohm meter (Agilent, 34420A) measures the voltage of device two ends and output to testing system for saving. The system loads pressure on sensor with different speeds and saves the pressure data. For sensitivity testing, the system gives a pressure to device with a speed of 1 mm/min until up to 150 kPa. In mechanical endurance testing, the system keeps a pressure range cycling from 1 kPa to 15 kPa on device with a speed of 6mm/min. To test response time, the system presses devices from 1 kPa with a 600 mm/min speed and stays 2 s after reaching 50 kPa. Since large changes of sensor resistance and testing system range limited, to show the trend clearer, it needs to convert voltage changing under the constant current to current changing under the constant voltage equivalently, which follows:

$$\frac{I}{I_0} = \frac{U_0}{U} \tag{1}$$

The voltage signal in range of (0, 1] can be converted to current ratio signal in range of [1, ∞) by this equation.

### 2.4. Preparation of Ion-Gel

The ion-gel is made from PVDF-HFP (Aladdin, CAS 9011-17-0) and [EMIM][TFSI] ion liquid (Energy Chemical, CAS 174899-82-2). The mixed liquid with a mass ratio of 1:7 of PVDF-HFP and acetone needs to stir 2 h in 50 °C silicone bath with 700 r/min and mixes it into ion liquid with a mass ratio of 1:4 of PVDF-HFP and [EMIM][TFSI] with the same procedure. After ultrasonically cleaning a glass slide at 25 kHz for 10 min twice and blowing dry with $N_2$, stirred mixed liquid will be dropped on it. After spin coating 90 s with 500r/min, the mixed liquid needs to place quickly into vacuum oven for drying 24 h in 70 °C vacuum environment.

### 2.5. Fabrication of Flexible Transistor Array

Flexible transistor array is made from PI substrate, ion-gel and Cr/Au base electrode. To assure flexibility, P3HT (Aladdin, CAS 104934-50-1) is used as semiconductor layer. 10mg/ml P3HT is mixed with ortho dichlorobenzene and stirred 2 h with 700 r/min in 50 °C silicone bath for dissolving. After P3HT and ion-gel preparation, little P3HT is dropped onto the channel and spin coated 30 s with 2000 r/min. After P3HT film 70 °C annealing 2h, ion-gel needs to cut into suitable size by laser cutting machine (Xtool, P2-55W) for covering part gate electrode and whole channel. This Ion-gel Gated Transistor (IGT) is prepared for array integration.

### 2.6. Characterization of Flexible Ion-Gate Transistor

IGT is characterized by semiconductor parameter analyzer (Keithley, 4200A-SCS) and cryogenic probe station (Lake Shore). 3 I-V source measure units (SMU) are connected to source, drain and gate respectively. Source electrode is loaded -0.05V and drain electrode is grounded. Applying different frequency pulses (3 Hz, 2 Hz, 1.4 Hz) on gate for testing spike frequency dependence plasticity. Applying different quantity pulses (10, 15, 20, 25) on gate for testing spike number dependence plasticity. Applying different duration pulses (200 ms, 300 ms, 400 ms, 500 ms) on gate for testing spike duration dependence plasticity. Applying different amplitude (-1.5 V, -2.0 V, -2.5 V, -3.0 V) on gate for testing spike amplitude dependence plasticity. Applying a bidirectional voltage sweeping from 0 V to 2.5 V on gate and a fixed voltage of -0.05 V on source for testing transfer characteristic. Applying a voltage sweeping from 0 V to -1.0 V with different gate voltage (-0.5V, -1.0 V, -1.5V, -2.0 V) for testing output characteristic.

## 2.7. Construction of Signal Processing Circuits

For input requirement of all devices, previous output signal needs to be modulated by signal process circuit. Resistance changing is converted to voltage signal in sensing circuit unit by bleeder circuit adopted. As negative piezoresistive effect, the output voltage of sensor is proportional to pressure input. Spike coding unit is designed for converting voltage signal to pulse signal. The relationship between voltage and pulse frequency ($f$) in spike-coding is following:

$$\frac{1000}{f} = 512 - 0.412 \left(\frac{1024 V_{in}}{5}\right) \tag{2}$$

Each spiking time interval influenced by voltage (Vin) changes the frequency. P3HT in transistor manufacture is a p-type organic semiconductor while ion-gel gate transistor needs a frequency input, therefore, a negative frequency input is required to transistor. An op-amp is adopted to scale the 5 V (high level of MCU) pulse signal to -2 V ($R_F/R$ = 8kΩ/20kΩ = 2/5) in inverting input mode. Drain current of transistor output from I/V converter in back-end as the input of 3 comparators (3 thresholds) and V1, V2, V3 as the output, a 2 bits integer output in area [0, 3] can be concatenated by 3 logic gate output (bit[0] = ($V_1$ XOR $V_2$) OR $V_3$, bit[1] = $V_2$ OR $V_3$). These 9 results from 3×3 array is arranged into an 18 bits integer data and sent to master machine in 2400 baud rates.

## 2.8. Algorithms for Signal Processing

Read the integer which contains 18bits binary number (9×2 bits integers) from COM (cluster communication) port, convert to string type value and get the maximum each 400 ms of COM signal stream (Note S1). Since the frequency minimum is 2.5 Hz, peak value exists. Two threads are used to process COM stream. COM stream is converted to 8 bits integer (2 bits blank interval from COM), combined 3×8 bits integer as a unit (an integer segment blank, 32 bits totally) and sent to data queue for another thread processing peak. Since 400ms time interval ($T$) and specified 2400 baud rates ($B$), each $N$ ($N$ = 30 in this work) integers can extract a maximum as:

$$N = \frac{TB}{1000 \times 32} \tag{3}$$

In these 30 integers adjacent difference values (current integer minus the previous), the quantity of negative value after the positive is the peak count in a time interval.

# 3. Results and discussion

*3.1. Design of Neuromorphic Tactile Architecture*

For high energy efficiency and universality, in this research, we design a neuromorphic tactile system by modular design including 3 independent parts corresponding to skin receptor, ganglion and central nervous system in biological tactile system respectively: sensor, synaptic transistor and neural network based on bionic design thinking. Through skin receptor in biological system, external signal is spike coded to accumulate for central nervous system recognizing and reacting (Fig. 1a, b, c). This procedure is fit for the process structure of sensing unit, front-end processing unit and back-end processing unit (38-40). With pressure as the information medium, we design a 3×3 sensor array as perceptional network sensing unit, a low power consumption MCU for spike coding and synaptic transistor as front-end processing unit (Fig. S1). The pressure signal from 3×3 sensor array will be converted to voltage signal by bleeder circuit for spike coding and modulating via low power MCU (Fig. 1d). The organic film transistor made from P3HT and ion-gel can simulate various synaptic plasticity for input signal preprocessing (Fig. 1e) (41). We design a circuit for peak value output and an ANN (Artificial Neural Network) with some algorithms for information reception and recognition as back-end unit (Fig. 1f). Just like toy bricks, this type of design makes each part of the system changeable and assembled and increases system compatibility.

*3.2. Design of Sensing Unit*

Skin can detect various input (pressure, pattern and temperature etc.) simultaneously via plenty of receptors, therefore, sensor unit need to be an array and 3D printing is a suitable technology for preparation. Not only sensitivity and range but also flexibility and producibility are important to tactile sensor unit. The sensing unit is a light curing 3D-printed conductive resin pressure sensitive layer (Fig. S2) which is packaged on polyimide (PI) membrane evaporated with Cr/Au interdigital electrodes (Fig. 2a, b). 3D-printed flexible pressure sensitive layer with a pyramidal microstructure for high performance will be deformed by loaded force, which make the interdigital electrodes in open circuit state connect to form an electric pathway. In this case, interelectrode resistance is decreasing with increasing pressure (Fig. 2c). The relationship between the sensitivity ($S$) in condition of a pair of interdigit and resistivity ($\rho$) with compressive displacement ($h$) is followed:

$$S(h, \rho) = \frac{R_0 h t}{\rho d_0 P} \tag{4}$$

$R_0$ is the resistance without force applying and $d_0$ is the distance between two fingers of interdigital electrodes. $t$ and $P$ denote electrode thickness and pressure respectively (Note S2). Less resistivity means higher sensitivity for any given compressing displacement. MWCNT can optimize the resin conductivity, therefore, we proposed a high performance Agilus 30/MWCNT/HDDA composite resin material. As Mechanical deterioration from mixed MWCNT with resin (Fig. S3), HDDA can alleviate this problem and improve the resistivity of resin. With HDDA concentration increasing, the range of 3D-printed sensor has been extremely extended (Fig. S4, 5). Finally, this material surface resistivity can reach only about 3.42 kΩ. The pressure sensitive sensor made from this material has a high sensitivity of 31.687 kPa$^{-1}$ and 11.712 kPa$^{-1}$ in pressure ranges [0, 4.85) kPa and [4.85, 36.28] kPa respectively, it even keeps 1.481 kPa$^{-1}$ sensitivity in condition of 150 kPa (Fig. 2d). Due to the rational design of exposure parameter (75 mW/mm$^2$, 1.5 s), our sensor only needs 90 ms to recover to its initial state and has a good stability after 500 cycling load-unload in mechanical endurance test (Fig. 2e, f). This sensor shows a good flexibility in performance test for bending ability (Fig. S6a). With a same produce procedure, in a test of 9 samples, the line charts of pressure-strain are almost identical (Fig. S6 B). The sensor has good uniformity and high sensitivity. Meanwhile, we tested the response of sensor in various frequency and pressure. As a result, our device shows good stability (Fig. S7).

The most sensors with voltage as output can be integrated in sensing unit of this system. For example, resistance (or capacity) of resistive (or capacitive) sensor can increase (positive effect) or decrease (negative effect) with increasing input signal. Bleeder circuits can provide a stable working environment of these 2 effect sensors for adapting the value of $R_{sensor}$, $R_{ref}$ and $C_{sensor}$, $C_{ref}$ (Fig.S8). Moreover, frequency signal can be also outputted directly by RC oscillating circuit. Since self-powered sensor limited by physical property, weaker signal emitting and dynamic signal monitoring only, an amplifier circuit is needed before spike coding for signal amplifying into a proper range. Here, a unit in pressure sensing array is replaced by commercial temperature sensor for universality testing. As a result, the system can output trend of temperature changing accurately (Video S1). Inspiring, we conceive a heterogeneous integration of multiple sensors to achieve system versatility (Fig. S9).

*3.3. Design of Front-end processing Unit*

Ion gate transistor with a high performance, bionic synaptic plasticity and preparation easily is suitable for front-end processing unit (42). P3HT and ion-gel provide a good flexibility of this organic transistor, moreover, array structure makes integrating more sensor units possible (Fig. 3a). Gate voltage can modulate the ion distribution in ion-gel via ionic transport to generate an electric double layer between electrolyte layer and semiconductor layer for reducing equivalent resistance between source and drain. In addition, with a sufficient frequency pulse, ions in ion-gel and holes in P3HT cannot be transported back totally to uniform distribution state in a pulse interval. Therefore, the next pulse can cause more holes to accumulate near electrolyte-P3HT interfaces to further decrease equivalent resistance between source and drain (Fid. 3b). These mechanisms allow the device to emulate LIF model.

Flat P3HT film and ion-gel film provide a reliable performance for valid current output with a tiny voltage between source and drain (Fig. 3c, d). The on/off current ratio of this transistor can reach 10$^4$ (Fig. 3e). Spiking number & rate dependent plasticity is the base of signal process mechanism we proposed. Drain current of transistor can be significantly modulated by frequency and number of input pulse (Fig. 3f, g). Meanwhile, synapse characteristics like spiking amplitude and spiking duration dependent plasticity are observed in this transistor to demonstrate its good bionic synaptic performance (Fig. S10). Since good uniformity, synaptic transistor array can be large-scale manufactured via spin-coating and laser cutting.

To process continuous signal like biological systems, a spike coding unit consisted of MCU and operational amplifier (Fig. S11a, b) is set for ion-gel transistor. Based on bionic synaptic plasticity, data can be preprocessed by front-end processing unit. Because of significantly affected by frequency and amplitude of the input signal, setting these two parameters properly for output of neuromorphic devices (e.g. ion-gel gate transistor and memristor) is very important. Since modulating input frequency in MCU mainly, MCU clock is 32 frequency division to reduce the power. As a result, MCU can still output 250k pulses per second at most theoretically. Combining with delay function, output frequency can be adapted in area (0, 250k]. As MCU in form of high (5 V) and low (0 V) level output, an op-amp is used for amplitude modulation. To fit for neuromorphic processing unit, a voltage amplitude in range of (0, 5$A_u$] V frequency signal can be outputted from feedback resistor ($A_u$ is amplification factor).

### 3.4. Design of Signal Processing

A signal processing strategy with the advantage of neuromorphic device for processing different signal input is needed before circuit (Fig. S11c, d) and ANN design. In our universal neuromorphic tactile perception system (Fig. 4a), we propose a signal processing solution based on peak count and peak value for tactile characteristic recognition. The signal will be converted to pulse signal after spiking encoding (pulse frequency has a positive relation to pressure, Fig. 4b, c) and accumulated to peak signal by bionic synaptic mechanism. What is more, the frequency of input signal is preserved during the procedure so that we can get frequency information by peak count in each fixed time interval.

This event-driven system can switch to sleep mode without pressure input. Peak count and peak value information will be collected via circuit each 400 ms (Fig. S12) for signal form, Morse code recognition and pressure pattern recognition. When the peak count changes continuously, the signal is judged to be a continuous signal and count value shows the pressure trend and relative magnitude (Fig. 4d). The signal should be a frequency signal if short gap exists in peak count information (Fig. 4e). One-threshold solution can meet a severe error in recognizing unstable signal fluctuating around threshold. Multi-threshold we proposed can give the solution and threshold information can deduce the changing of signal. A detection limit is set for valid information capture in multi-threshold strategy. By the multi-threshold strategy and cumulative effect, signal output from synaptic transistor can reach different thresholds and be converted to the quantity value of threshold it exceeds (Fig. 4f, g). Therefore, threshold information demonstrates how much and how long the pressure applied. Threshold information including temporal information, so recognition task can significantly decrease data processing of back-end unit with a high accuracy.

Neural network type in this system and threshold quantity can be altered in back-end. Limited by technic, only 3 thresholds are set for comparator in our system, although more thresholds can be beneficial for recognition and combined to multi-state memristor. Here, what a more reliable scenario we conceived is transistor threshold voltage to be the threshold and integrated with logic circuit following. It is facilitated for threshold circuit to integrate with logic circuit. With excellent compatibility of neural network, ANN program we coded can construct a simple network like FCNN or CNN for data training and deducing. For more details, please retrieve it on our [GitHub repository](#).

*3.5. Tactile Characteristic Recognition*

For verification of diverse system functionality, various pressure inputs are used for tactile perceptional system recognition (Fig. 5a). When a continuous changing signal with rising-holding-falling trend applies to the sensor array, the pressure signal is converted to pulse signal. Pressure changing information from pressure signal is included in electrical pulse signal (Fig. S13). After accumulating in synaptic transistor, peak value and peak count information are included in output (Fig. S14). Here, peak count information is used for signal form classification. Peak count information of continuous changing signal shows the same rising-holding-falling trend as pressure input (Fig. 5b). Likewise, peak count information of frequency signal also shows the same format of 3 short segment as pressure input (Fig. 5c). Different frequency of signal represents its own meaning such as the permutation of dot & dash in Morse code. We choose 5 letters "N", "I", "M", "T" and "E" as Morse code to map pressure action time from signal input (Fig. 5d, e). To differentiate the long, short and continuous signal, peak count is the key to solution. Here, we stipulate that count value in range [10, 20) is dash signal and [0, 10) is dot signal. When the count value is over 20, the input signal will be considered as a continuous signal. We design a method to collect each signal segment peak count and interpret the information from different frequency signal input (Fig. S15). After several tests, we find that peak count value 10 is appropriate as a boundary to distinguish dash and dot signal. According to this rule, Morse code signals are recognized effectively (Fig. 5f). As a result, the recognition accuracy of letter "N", "I", "M", "T" and "E" can reach 90%. We invited five people to tap the Morse code of these five letters, 10 times per letter, and plotted the result. The confusion matrix shows that neuromorphic tactile system has a high accuracy in Morse code recognition (Fig. S16).

Different from peak count information, peak value information is a tool of recognizing pressure pattern from tactile sensor unit. To avoid ADC in circuit design, 3 thresholds are mapped to 3 comparators. Circuit will output the number of the threshold which the analogue signal maximum surpassed. What is more, accumulated value of signal includes temporal information due to the cumulative effect of neuromorphic device so that maximum peak is sufficient for ANN recognizing accurately (Fig. 6a to 6c). Four basic operator handwriting symbols with different writing habits "+", "−", "×" and "/" are designed to wooden molds (Fig. 6d). After mold gives pressure to tactile sensor unit, circuit converts the signal to a 1×9 column vector and sent to a 5 layers ANN for recognition (Fig. 6e, Fig. S17) and reach an accuracy of 87% and recall rate of 80% (Fig. 6f). In addition, deducing accuracy can be 90% in inference (Fig. 6g). Compared to normal procedure, power consumption of the neuromorphic system and bionic signal processing solution in accurately recognizing the same data can reduce 90.46% (Fig. 6h, Note S3).

## 4. Conclusion

We successfully designed a universal fully flexible near-sensor neuromorphic tactile perception system. By modular design, the sensing unit, neuromorphic device and back-end neural network are designed separately. Simple circuits in this system can distinguish the signal type. These features make system extremely extensible and compatible for updating or adapting. Because of the capability of forming EDL (electrical double layer) in metal gate & ion-gel interface of IGT, channel resistance is affected by ion distribution in ion-gel. This feature allows the device to mimic synaptic plasticity and generate peak data according to the input pulses. Peak information extraction and utilization significantly reduce system data processing. Meanwhile, multi-threshold signal processing strategy of our system allows diverse input. The system can get the trend of continuous or frequency signal and reach an accuracy of 90% and 87% in Morse code recognition and pressure pattern recognition respectively. Compared to conventional signal processing strategies, this event-driven system reduces 90.46% power consumption by using pulse signal and working only in waken thread for nonvoid data processing. This system provides a new path with great development prospects and applied value for intelligent robotics, brain-inspired computing and other fields.

## Conflict of interest

The authors declare that they have no conflict of interest.

## Acknowledgments

This research was partially supported by National Natural Science Foundation of China (U22A20248, U24A6001, 52127803, U24A20228, U22A2075, 62174165, 52301256, 52401257, 52201236, M-0152), National Key R&D Program of China (2024YFB3814100, 2023YFC3603500), International Partnership Program of Chinese Academy of Sciences(181GJHZ2024138GC), Chinese Academy of Sciences Youth Innovation Promotion Association (2018334), Talent Plan of Shanghai Branch, Chinese Academy of Sciences (CASSHB-QNPD-2023-022), Project of Zhejiang Province(2022R52004), Ningbo Technology Project (2022A-007-C), Ningbo Natural Science Foundations (2022J288, 2023J049, 2023J345), Ningbo Key Research and Development Program (2023Z097, 2024Z148, 2024Z143, 2024Z199, 2024Z171), Ningbo Public Welfare Program (2023S067).

## Author contributions

L.J. and L.D. designed the experiments. L.J., X.X. and W.Y. prepared the 3D-printed sensor array. L.J., L.X., Z.X. and J.W. prepared the synaptic transistor. L.J. and T.H. designed and fabricated FPC circuits. L.J., L.D., L.Z. and H.Z. designed and implemented the whole back-end unit. L.Y., W.Y. and B.B. helped with data analysis. L.J. and L.D. prepared the paper. S.J. and L.R.-W. supervised the research. All authors discussed the data and revised the text.

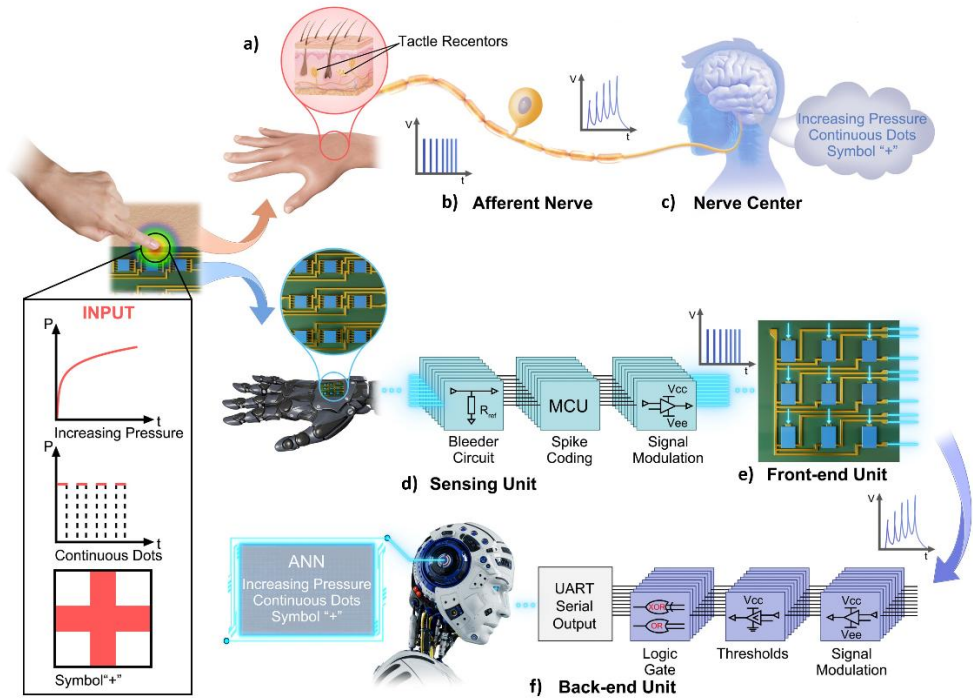

**Fig. 1. Design of bio-inspired neuromorphic tactile perception system.** (**a**) Tactile receptor, including Pacinian corpuscles and Merkel disks, is widely distributed in the dermis. It can convert external stimuli to nerve impulses. (**b**) Tactile afferent nerve pathway with ganglion. Here, nerve impulses are preprocessed by ganglion and then sent to neural center. (**c**) The primary function of nerve center is input information identification and reaction along with the received information. (**d**) To simulate the functionality of tactile receptor, flexible pressure sensor array and spike coding circuit are used to convert signal from pressure to pulse. A sensing unit corresponds to a tactile receptor. (**e**) By synaptic plasticity, the 3×3 synaptic transistor array is used to preprocess input pulse signal. (**f**) Back-end processing unit can be subdivided into two parts. One is to extract peak and threshold information by simple circuits and algorithms. Another is to identify and recognize input information by FCNN (Fully Connected Neural Network).

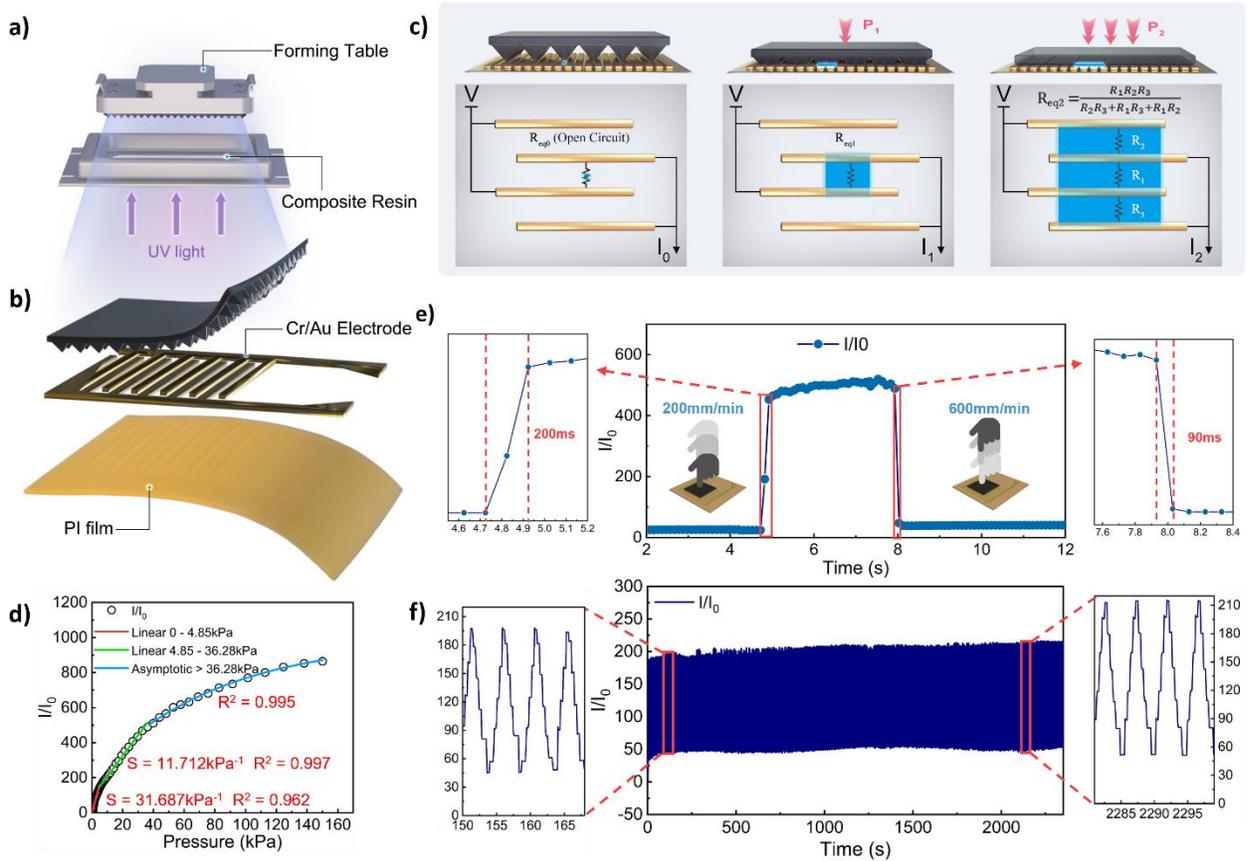

**Fig. 2. Design and Characterization of flexible 3D-printed sensor.** (**a**) Work mechanism of DLP 3D printer. By defining the irradiation area, the light-curing resin is cured into a variety of shapes layer by layer on the forming table with a 405 nm UV light. (**b**) Sketch map of flexible 3D-printed sensor. The sensor comprises a conductive pressure sensitive layer, Cr/Au interdigital electrode and PI substrate, arranged in a sequence from upper to lower. (**c**) Work mechanism of pressure sensitive sensor. With the increasing pressure, the contact area between interdigital electrodes and conductive layer expands, while the resistance between electrodes decreases. (**d**) Current ratio-pressure curve of 3D-printed sensor. The sensitivity of sensor can reach 31.687kPa$^{-1}$ below 4.85kPa and keep more than 1kPa$^{-1}$ over 150kPa. (**e**) Response time of 3D-printed sensor with 50kPa pressure loaded and pressing & releasing speed of 200mm/min and 600mm/min respectively. (**f**) Mechanical endurance of 3D-printed sensor with 15kPa pressure loaded.

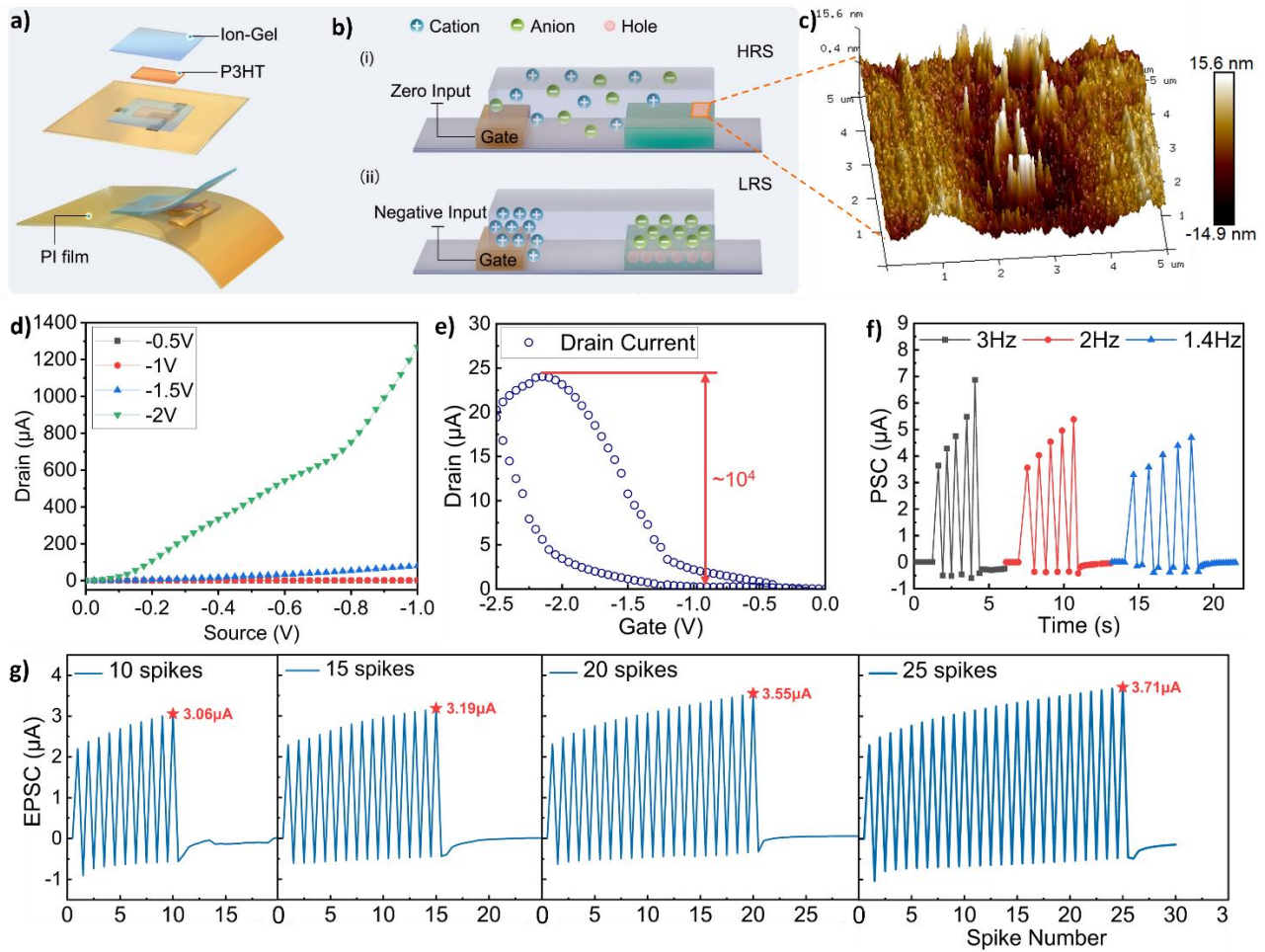

**Fig. 3. Design and Characterization of flexible synaptic transistor.** (**a**) Sketch map of flexible synaptic transistor. The transistor is made from ion-gel film, organic semi-conductor film, Cr/Au electrode and PI substrate, arranged in a sequence from upper to lower. (**b**) Working mechanism of synaptic transistor. (i) When gate is no voltage loaded, anion and cation are distributed uniformly throughout the ion-gel, resulting in a high resistance state (HRS) in the channel. (ii) When a negative voltage is loaded, anion and cation will accumulate at electrolyte/channel interface and electrode/electrolyte interface respectively, resulting in a low resistance state (LRS) in the channel. (**c**) Macroscopic image and SPM image of P3HT synaptic transistor channel. The surface of P3HT film has a low roughness. (**d**) Output characteristic curve of synaptic transistor. (**e**) Transfer Characteristics of synaptic transistor. Gate voltage sweeps from 0V to 2.5V with a source voltage of -0.05V. (**f**) Spiking rate dependent plasticity of the synaptic transistor with intervals of 300ms, 500ms and 700ms respectively. (**g**) Spiking number dependent plasticity (SNDP) of the synaptic transistor with 10, 15, 20, 25 spikes input.

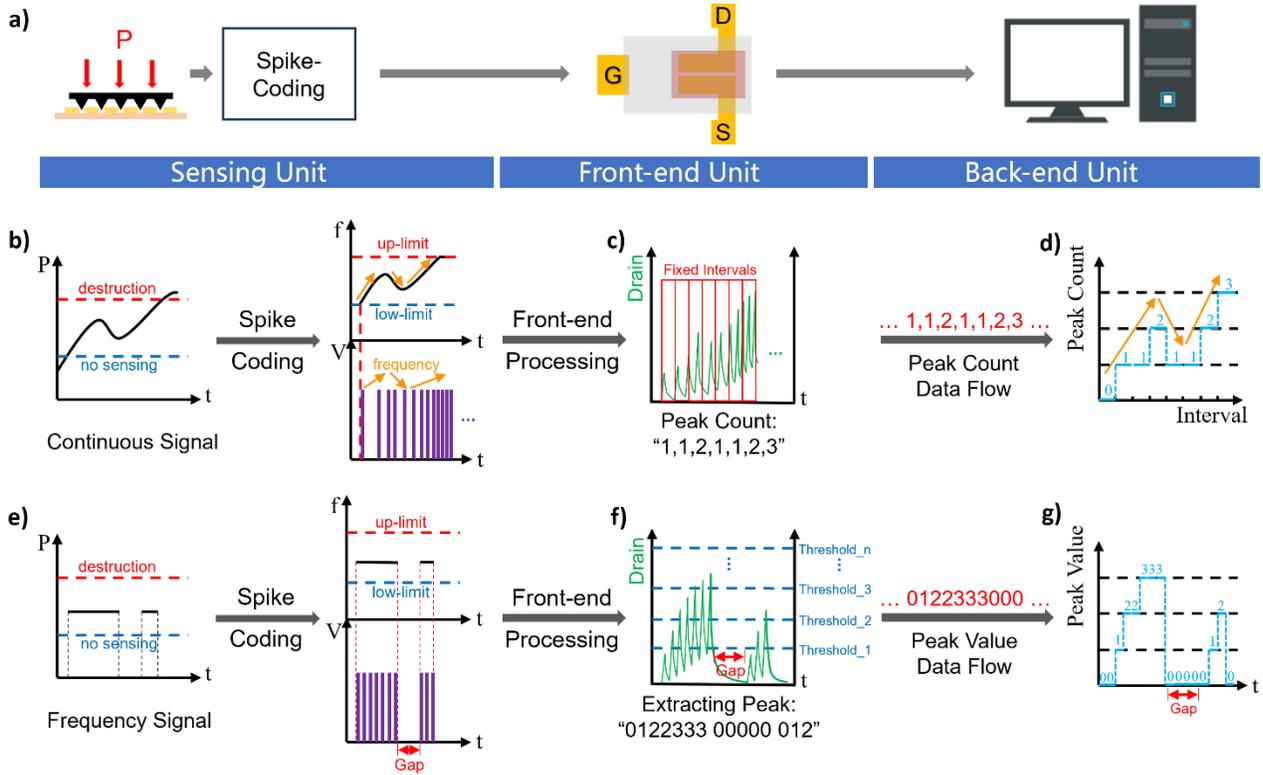

**Fig. 4. Signal Process mechanism of neuromorphic tactile perception system.** (**a**) The neuromorphic tactile system has 3 parts: sensing unit, front-end processing unit and back-end processing unit. (**b**) Input and output of sensing unit. After spike coding continuous signal, input pressure is converted to frequency signal because of the positive relationship between force and high frequency signal quantity. (**c**) Synaptic transistor corresponding output. Peak count output is decided by frequency of the input signal, which can show the changes of pressure directly. (**d**) For continuous changing signal, peak count in fixed interval corresponds to the trend of pressure correctly. (**e**) Input and output of a group of long and short signals. Frequency signal is not continuous with a gap in it. (**f**) Corresponding output of synaptic transistor should also be not continuous with a gap in it. (**g**) For peak value information, the highest peak can reflect the pressure loading duration or pressure level.

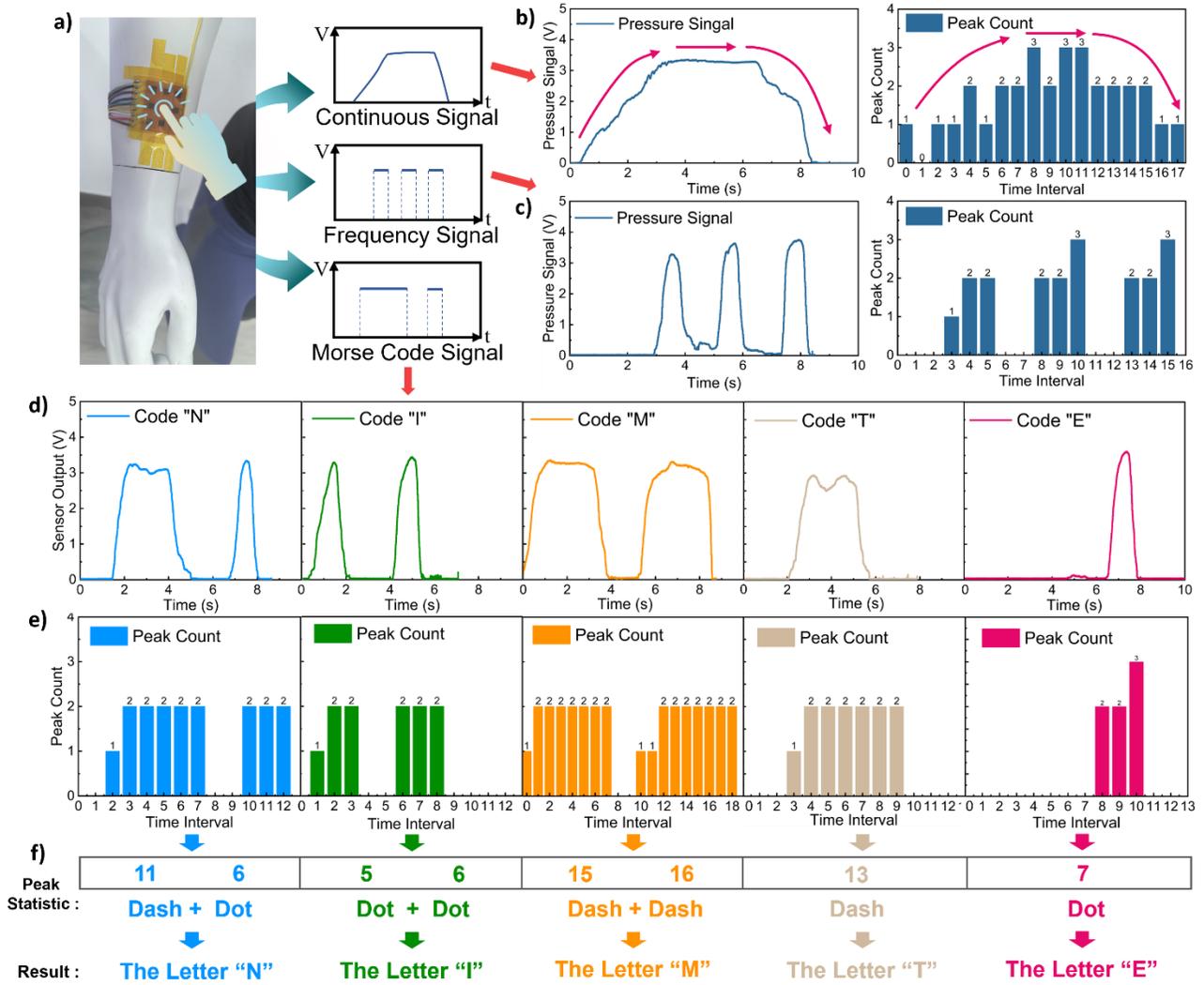

**Fig. 5. Signal discrimination and Morse code recognition by neuromorphic tactile perception system.** (**a**) Photograph of neuromorphic tactile system stuck on the artificial hand. (**b**) Continuous signal input and corresponding peak count output in each 400ms of system. Both input and output signal have a rising-holding-falling trend. (**c**) Frequency signal input and corresponding peak count output of system. Both input and output signal have 3 short segments. (**d**) Pressure signal of Morse code "N", "I", "M", "T" and "E" (the Morse codes of these five letters are "–•", "••", "– –", "–" and "•", respectively). (**e**) Peak count output in each 400ms of five letters. (**f**) Recognition results of five letters. Based on the proposed rule (i.e., when peak count < 10, the signal is recognized as dot, otherwise dash or continuous signal), peak count can be transferred to dash & dot combination (Video S2).

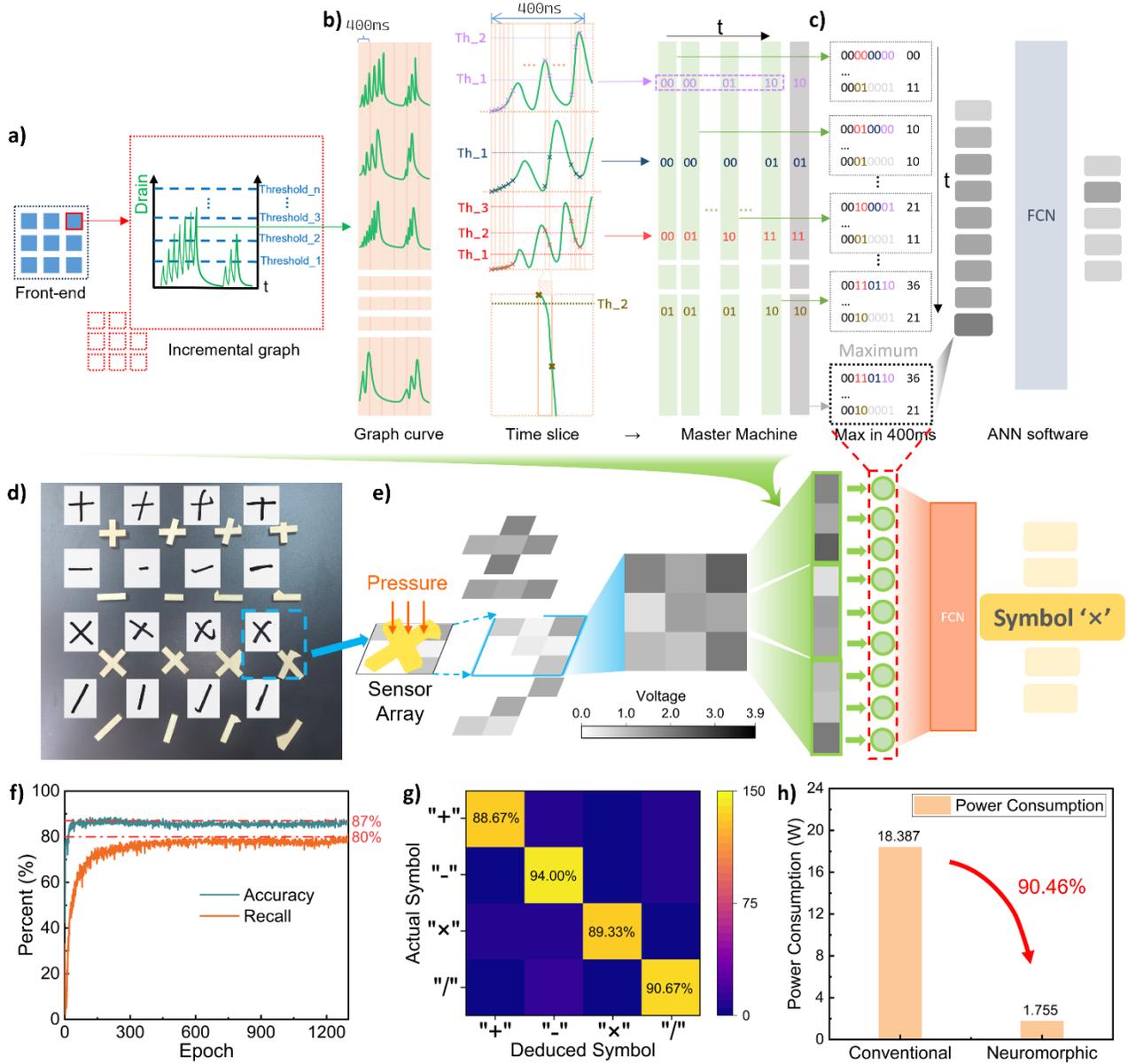

**Fig. 6. Tactile Pattern recognition by neuromorphic tactile perception system.** (**a**) Accumulated output signal from front-end processing unit. (**b**) UART serial communication protocol of system. (**c**) Maximum in each 400ms is extracted as input of FCNN for pattern recognition. (**d**) Handwriting mathematical operator sample. (**e**) Sketch map of pressure signal converting. The 3×3 in-put pressure signal array is arranged to a 9×1 column vector as the input of neural network. (**f**) Curve of accuracy and recall of FCNN. It can reach 87% accuracy and 80% recall in training. (**g**) Recognition results of 4 basic mathematical operators (Video S3). The confusion matrix shows that FCNN is suitable for tactile pattern recognition with an accuracy of about 90%. (**h**) Power consumption of conventional tactile system and neuromorphic tactile system is 18.387W and 1.755W respectively. Obviously, power consumption decreased 90.46%.

18